\documentclass{article}

\usepackage{spconf}
\usepackage{amsmath,graphicx}
\usepackage{enumitem}
\usepackage{mwe}
\usepackage{cite}
\usepackage{booktabs}
\usepackage{makecell}
\usepackage{multirow}
\usepackage{graphicx}
\usepackage{hyperref}
\usepackage{newtxtext}
\usepackage{newtxmath}
\usepackage{csquotes}
\MakeOuterQuote{"}

\title{
3D-LLDM: LABEL-GUIDED 3D LATENT DIFFUSION MODEL \\ 
FOR IMPROVING HIGH-RESOLUTION SYNTHETIC MR IMAGING \\ 
IN HEPATIC STRUCTURE SEGMENTATION
}

\name{%
\begin{tabular}{@{}c@{}}
Kyeonghun Kim$^{1}$, Jaehyeok Bae$^{2}$, Youngung Han$^{1, 3}$, Joo Young Bae$^{3}$, Seoyoung Ju$^{1, 4}$, Junsu Lim$^{1, 4}$, \\
Gyeongmin Kim$^{5}$, Nam-Joon Kim$^{3, \dagger}$, Woo Kyoung Jeong$^{6}$, Ken Ying-Kai Liao$^{7}$, \\
Won Jae Lee$^{6, 8}$, Pa Hong$^{9}$, and Hyuk-Jae Lee$^{3}$
\end{tabular}
}

\address{
    $^{1}$ OUTTA, Republic of Korea \quad 
    $^{2}$ Stanford University, USA \\ 
    $^{3}$ Seoul National University, Republic of Korea \quad 
    $^{4}$ Sangmyung University, Republic of Korea \\ 
    $^{5}$ Chung-Ang University, Republic of Korea \quad 
    $^{6}$ Samsung Medical Center, Republic of Korea \\ 
    $^{7}$ NVIDIA, Taiwan \quad 
    $^{8}$ Sungkyunkwan University School of Medicine, Republic of Korea \\ 
    $^{9}$ Samsung Changwon Hospital, Republic of Korea
}

\begin{document}
\ninept 

\setlength{\abovedisplayskip}{3pt}
\setlength{\belowdisplayskip}{3pt}
\setlength{\abovedisplayshortskip}{2pt}
\setlength{\belowdisplayshortskip}{2pt}

\setlength{\textfloatsep}{8pt plus 1.0pt minus 2.0pt}
\setlength{\floatsep}{8pt plus 1.0pt minus 2.0pt}
\setlength{\intextsep}{8pt plus 1.0pt minus 2.0pt}

\setlength{\abovecaptionskip}{1pt}
\setlength{\belowcaptionskip}{0pt}

\maketitle

\begin{abstract}
Deep learning and generative models are advancing rapidly, with synthetic data increasingly being integrated into training pipelines for downstream analysis tasks. However, in medical imaging, their adoption remains constrained by the scarcity of reliable annotated datasets. To address this limitation, we propose 3D-LLDM, a label-guided 3D latent diffusion model that generates high-quality synthetic magnetic resonance (MR) volumes with corresponding anatomical segmentation masks. Our approach utilizes hepatobiliary phase MR images enhanced with Gd-EOB-DTPA contrast agent to derive structural masks for liver, portal vein, hepatic vein, and hepatocellular carcinoma, which subsequently guide volumetric synthesis through a ControlNet-based architecture. Trained on 720 real clinical hepatobiliary phase MR scans from Samsung Medical Center, 3D-LLDM achieves Fréchet Inception Distance (FID) of 28.31 and a 70.9\% improvement over GANs and 26.7\% over state-of-the-art diffusion baselines. When used for data augmentation, our synthetic volumes boost hepatocellular carcinoma segmentation by up to 11.153\% Dice score across five CNN architectures. 
\end{abstract}

\vspace{-0.1cm}
\section{Introduction}
\vspace{-0.1cm}
Deep learning has achieved remarkable advancements in extracting relevant patterns from data and making precise decisions, particularly in image-based tasks such as classification and segmentation \cite{a1cai2020, a20minaee2021image, a25hao2019review, a18xiaoqing2021advances}. The performance of deep learning models has rapidly improved, frequently surpassing traditional methods across various domains.

Generative models are developed to capture underlying data distributions
and produce new, synthetic images. Recent studies have shown that traditional generative models can be applied to the medical domain, including MR and CT imaging \cite{a10jha2024ct,a13kang2021synthetic,a24shin2021deepgan}. However, their ability to synthesize high-quality medical
volumes remains insufficient for practical applications.

\begin{figure}[t]
\centering
\makebox[\columnwidth]{\includegraphics[width=1.01\columnwidth]{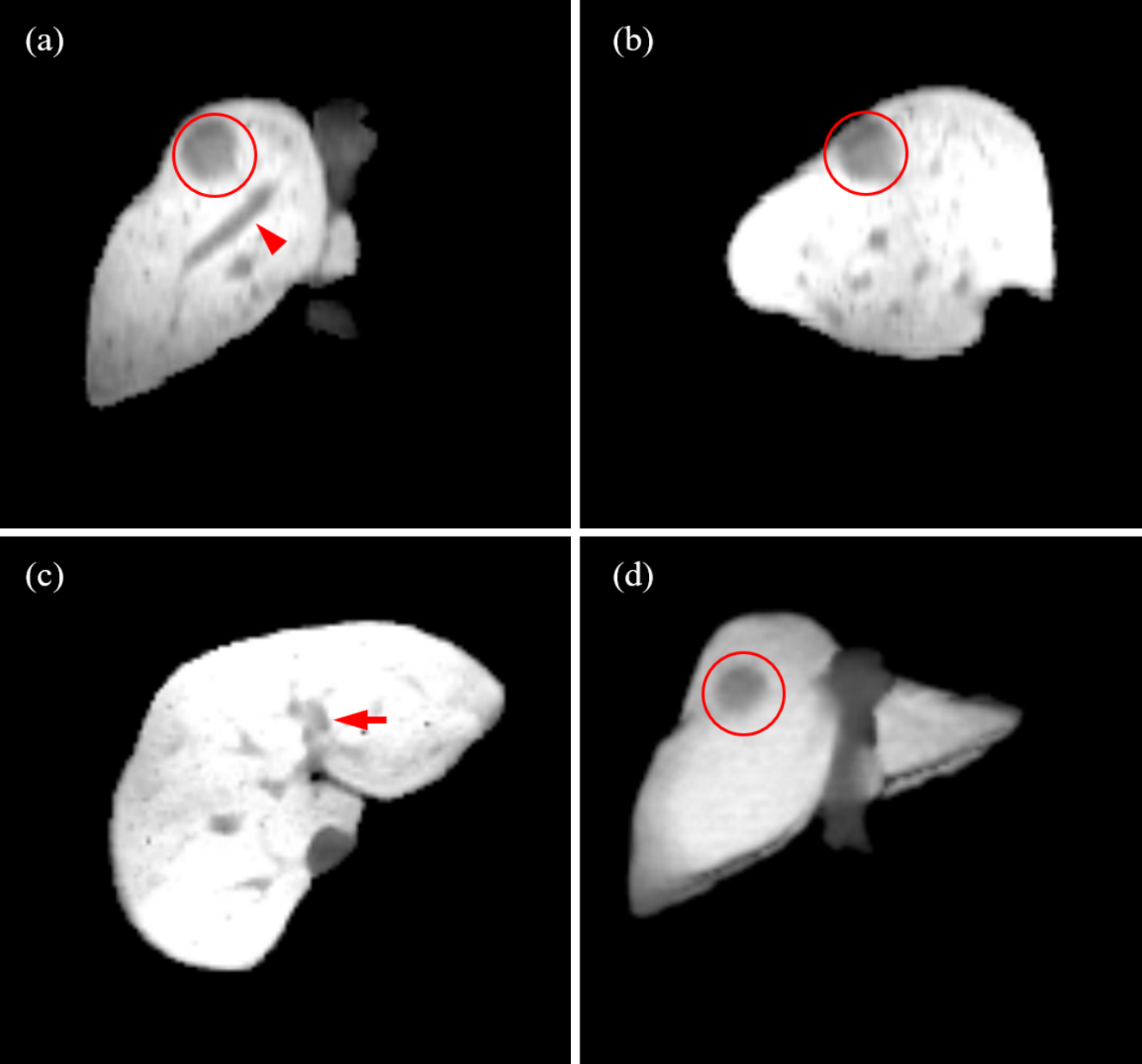}}
\vspace{-6mm}
\caption{
High-Quality synthetic MR volume generated by 3D-LLDM demonstrating anotomical consistency across multi planes: (a) Coronal image, (b) sagittal image, and (c) axial image reformatted from the
high-resolution 3D volumetric data shown in (d). Note the tumor highlighted by a red circle, the right hepatic vein indicated by an arrowhead, and the left portal vein marked by an arrow.}
\label{fig1}
\vspace{-1mm}
\end{figure}

A major challenge in this field is the limited availability of high-quality training data. Since deep learning models rely on extensive and diverse datasets for effective training, the shortage of high-quality medical images restricts the accuracy and realism of the generated synthetic volumes \cite{a12kang2023label,a17liu2023clip,a30zhou2019high}.

To overcome this constraint, we introduce an innovative approach --- the Label Guided 3D Latent Diffusion Model (3D-LLDM) --- designed to generate high-resolution synthetic volumes along with their corresponding segmentation labels.
In the context of abdominal MRI, our model first synthesizes segmentation labels containing the liver and veins, which are then used to train ControlNet \cite{a27controlnet} as spatial
guidance for volume generation. Leveraging these guided synthetic labels, our model produces realistic synthetic datasets, which can be effectively utilized for downstream tasks. 

Figure~\ref{fig1} shows coronal, sagittal, and axial views
reformatted from 3D volumetric data, highlighting hepatic
veins and a tumor. This example illustrates the importance
of generating anatomically consistent 3D synthetic volumes.

Our model is trained with the HBP images of 720 Gd-EOB-DTPA-enhanced MR imaging and generates better high-resolution synthetic images compared to conventional 2D GAN-based models \cite{a11hagan}.  Additionally, the synthesized volume-label pairs are incorporated into the training datasets for downstream tasks of two different segmentation methods of the hepatic structures including the liver segmentation and multi-class segmentation from the portal vein, hepatic vein and tumor (i.e., hepatocellular carcinoma [HCC]). Comparative results among various CNN-based segmentation models \cite{a22ronneberger2015U-net, a29zhang2018road} demonstrate that the high-resolution synthetic volumes generated by 3D-LLDM significantly improved the performance of segmentation of the hepatic structures. 

Our main contributions are: (1) to present 3D-LLDM, an innovative diffusion based generative model tailored for high-resolution MR volume synthesis, incorporating label guidance to improve generation control and (2) to demonstrate that incorporating 3D-LLDM-generated synthetic data into the training pipeline could improve the performance of CNN-based liver segmentation models, achieving higher accuracy compared to training on real data alone.
\vspace{-0.1cm}
\section{Related Work}
\vspace{-0.1cm}
\subsection{Diffusion Model}
\vspace{-0.2cm}
Diffusion-based models have emerged as a powerful generative approach, widely applied in medical imaging tasks such as image synthesis \cite{a9hung2023med, a23Shi2025diffusion, a14kazerouni2023diffusion, a7DDPM, a4croitoru2023diffusion} due to
their capabilities in high-quality image synthesis. Unlike previous approaches, these models excel in generating diagnostically valuable 2D medical images with remarkable detail fidelity. However, extending to 3D remains challenging; current methods like text-guided CT generators create volumes slice-by-slice, resulting in problematic structural inconsistencies between cross sections \cite{a5hamamci2024generatect}. While a recent study proposed a slice-based LDM for MRI synthesis \cite{kebaili20243d}, it may lack full volumetric coherence. In contrast, our 3D-LLDM operates directly in 3D latent space with label-guided ControlNet, ensuring strict anatomical consistency across all planes.
Research has increasingly targeted specialized applications, particularly synthetic tumor generation for improving segmentation performance across multiple organs \cite{a8hu2022synthetic, a16lai2024pixel, a19lyu2022learning, a26wu2024freetumor, a28zhang2023self}.

\vspace{-0.2cm}
\subsection{3D Latent Diffusion Model}
\vspace{-0.2cm}
Latent Diffusion Models (LDMs), a variant of diffusion model, operate in a compressed latent space, significantly reducing computational complexity while
preserving high-quality outputs \cite{a21LDM}. 

Given a medical volume $x \in \mathbb{R}^{H \times W \times D}$, where H, W, and D represent the dimensions of height, width, and depth, respectively, the encoder $\mathcal{E}$ maps $x$ to a latent representation $z = \mathcal{E}(x) \in \mathbb{R}^{h \times w \times d}$. The decoder $\mathcal{D}$ then reconstructs the volume as $\tilde{x} = \mathcal{D}(z)$ from the latent features. Typically, a Variational Autoencoder (VAE) structure \cite{a15vae} is employed for the encoder-decoder pair $(\mathcal{E}, \mathcal{D})$.

In Latent Diffusion Models (LDMs), the model $\epsilon_\theta(z_t, t)$ is trained to estimate 
a noise-free reconstruction of the input latent features $z_t$, where $z_t$ represents a
noisy transformation of the original input at time step $t \in [0, T]$. The training objective for the latent diffusion model $\epsilon_\theta$ is formulated as:
\begin{equation}
\mathbb{E}_{\mathcal{E}(x), \epsilon \sim
\mathcal{N}(0, 1), t} \left[ ||\epsilon - \epsilon_{\theta}(z_t, t)||_2^2 \right].
\end{equation}

\subsection{ControlNet}
\vspace{-0.2cm}
ControlNet \cite{a27controlnet} is a neural network extension that enhances diffusion models by incorporating additional spatial guidance, allowing precise control over the generated output through structured inputs. Combining with a Latent Diffusion Model (LDM), the additional conditioning information is transformed into latent features, denoted as the task-specific condition $c_f$. The overall learning objective of the diffusion algorithm, incorporating ControlNet $\epsilon_\theta$, is formulated as:
\begin{equation}
\mathbb{E}_{\mathcal{E}(x), \epsilon \sim
\mathcal{N}(0, 1), t, c_f} \left[ ||\epsilon - \epsilon_{\theta}(z_t, t, c_f)||_2^2 \right].
\end{equation}

In this work, we leverage the NVIDIA MONAI framework an open-source AI platform tailored for medical research \cite{a2cardoso2022monai} to effectively adapt ControlNet for medical image processing.

\vspace{-0.1cm}
\section{Methods}
\vspace{-0.1cm}
\subsection{Label-Guided 3D Latent Diffusion Model}
\vspace{-0.2cm}
In this paper, we propose the Label-Guided 3D Latent Diffusion Model (3D-LLDM) for generating high-resolution synthetic medical volumes paired with corresponding labels. Our approach first trains a synthetic label generation model using a latent diffusion framework. Given the encoder-decoder pair $(\mathcal{E}_l, \mathcal{D}_l)$ for label synthesis, the diffusion model $\epsilon_{l,\theta}$ is optimized using the following learning objective:
\begin{equation}
\mathcal{L}_{\text{diffusion}} = \mathbb{E}_{\mathcal{E}_l(l),\, \epsilon_l \sim \mathcal{N}(0,1),\, t} \left[ \left\| \epsilon_l - \epsilon_{l, \theta}(z_l, t, t) \right\|_2^2 \right].
\end{equation}

\begin{figure}[t]
\centering
\vspace{-3mm}
\makebox[\columnwidth]{\includegraphics[width=1.01\columnwidth]{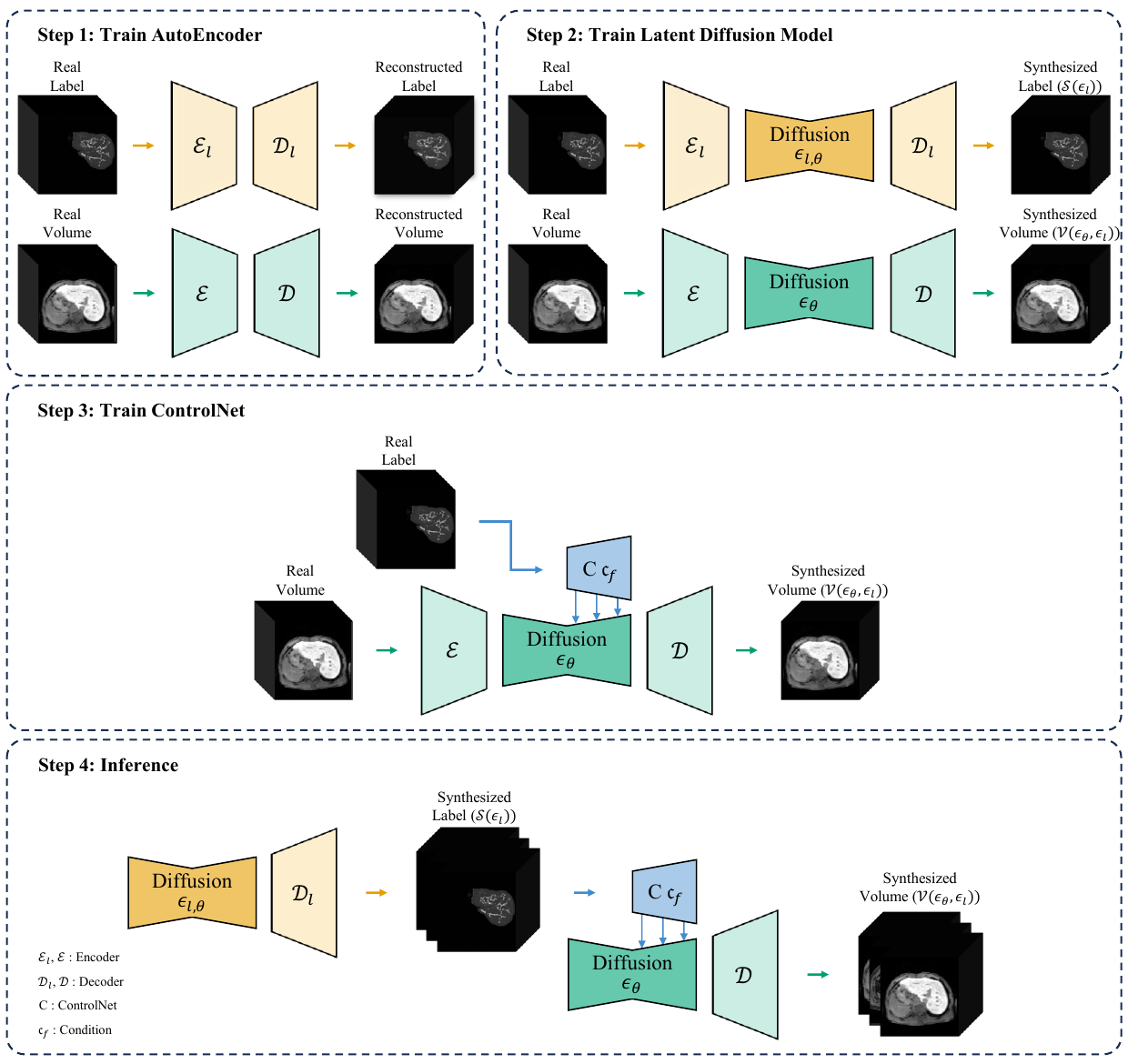}}
\vspace{-7mm}
\caption{
Overview of the training and inference pipeline of the proposed 3D-LLDM.
The process consists of four steps: (1) training the autoencoder for both label and volume reconstruction,
(2) training the latent diffusion model for label and volume synthesis,
(3) training the ControlNet to condition the diffusion process on labels,
and (4) performing inference using the pretrained latent diffusion models and ControlNet.
}
\label{overview}
\vspace{-1mm}
\end{figure}

Next, 3D-LLDM utilizes the synthetic labels 
$L(\epsilon_l) = \mathcal{D}_l(\epsilon_{l,\theta}(\epsilon_l, T))$ 
as input to ControlNet, which guides the generation of medical volumes. The task-specific condition $c_f(\epsilon_l)$ in the latent space of ControlNet is defined as:
\begin{equation}
c_f(\epsilon_l) = \mathcal{E}\!\left( \mathcal{D}_l(\epsilon_{l,\theta}(\epsilon_l, T)) \right).
\end{equation}

Then, the learning objective of ControlNet $\epsilon_{\theta}$ is formulated as:
\begin{equation}
\mathcal{L}_{c}=\mathbb{E}_{\mathcal{E}(x), \{\epsilon, \epsilon_l\} \sim
\mathcal{N}(0, 1), t} \left[ ||\epsilon - \epsilon_{\theta}(z_t, t, c_f(\epsilon_l))||_2^2 \right].
\end{equation}

In practice,  we use the latent-space representations of real labels,\(z_l\), instead of synthetic labels to improve performance, as shown in the following equation:
\begin{equation}
\mathcal{L}_{c,\text{train}} = 
\mathbb{E}_{\mathcal{E}(x), \{\epsilon_c\} \sim \mathcal{N}(0,1),\, t}
\left[ \left\| \epsilon - \epsilon_\theta(z_t, t, z_l) \right\|_2^2 \right].
\end{equation}

\begin{figure}[t]
  \centering
  \makebox[\columnwidth]{\includegraphics[width=1.01\columnwidth]{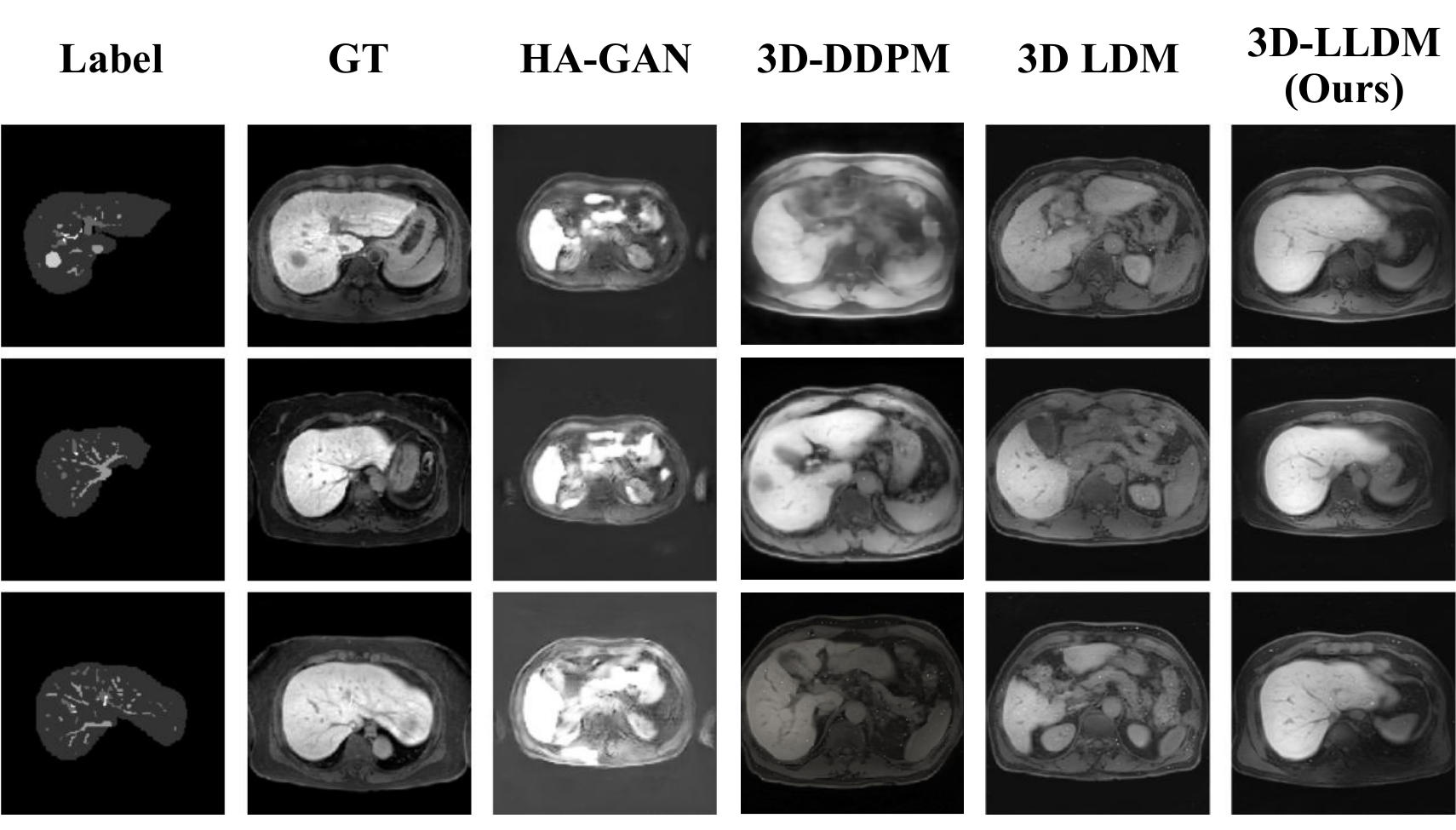}}
  \vspace{-7mm}
   \caption{Qualitative comparison of synthetic MR volumes of various levels generated by state-of-the-art generative models. The columns correspond to: (1) the label-guided input, (2) the ground truth (GT) MRI scan, (3) the HA-GAN-generated synthetic image, (4) the 3D-DDPM-generated synthetic image, (5) the 3D Latent Diffusion Model (3D LDM) with an autoencoder (AE) backbone, and (6) the proposed 3D-LLDM (Ours)-generated synthetic image.}
  \label{fig:onecol}
\end{figure}

Figure~\ref{overview} presents the structure and operational process of the proposed 3D-LLDM. The training process consists of two steps. First, our approach trains the autoencoder pairs $(\mathcal{E}_l, \mathcal{D}_l)$ for label synthesis and $(\mathcal{E}, \mathcal{D})$ for volume reconstruction. Subsequently, we train the latent diffusion model $\epsilon_{l, \theta}$ to generate synthetic labels, and ControlNet $\epsilon_{\theta}$ to generate synthetic volumes conditioned on the real labels. 

The inference process is illustrated in Step 3. Given random noise $\epsilon_{l} \sim \mathcal{N}(0, 1)$, we first generate synthetic labels as:  
\begin{equation}
S(\epsilon_l) = \mathcal{D}_l(\epsilon_{l, \theta}(\epsilon_l, T)).
\end{equation}

Next, using the generated synthetic labels and additional noise $\epsilon \sim \mathcal{N}(0, 1)$, we synthesize the final medical volume:
\begin{equation}
V(\epsilon_{\theta}, \epsilon_l) = \mathcal{D}(\epsilon_{\theta}(z_t, t, c_f(\epsilon_l))).
\end{equation}

This process produces a paired dataset of synthetic labels $S(\epsilon_l)$ and corresponding synthetic volumes $V(\epsilon_{\theta}, \epsilon_l)$, which can be effectively leveraged for downstream tasks such as segmentation.

\vspace{-0.1cm}
\section{Experiments}
\vspace{-0.1cm}
\subsection{Dataset and Implementation Details}
\vspace{-0.2cm}
This study employs anonymized MR imaging using Gd-EOB-DTPA (Bayer HealthCare, Germany) in 720 patients with HCC from Samsung Medical Center (Seoul, Korea). Particularly, we focused on volumes from HBP images, where the hyperintense liver is well discriminated compared to other hypointense abdominal organs. For each patient, we created segmentation labels encompassing the liver, portal vein, hepatic vein, and tumor (i.e., HCC), which serve as the ground truth for segmentation tasks.

To facilitate efficient training and enhance patient privacy protection, we
cropped the region of interest (ROI) to a fixed size of $(160, 160, 64)$ before
training. After cropping, the data was normalized via range scaling, mapping
intensity values to $[0, 1]$.

The dataset was divided into three subsets: 504 patients were designated for model training, 72 for validation purposes, and 144 for final testing evaluation. All computational experiments were performed using a single NVIDIA A100 GPU with 80GB memory, utilizing the MONAI 1.10 framework and Python 3.8 environment. The AdamW optimizer from PyTorch was employed for all gradient-based optimization procedures. The complete training phase for the 3D-LLDM required approximately one week of continuous computation on the A100 hardware.



\begin{table}[t]
\centering
\footnotesize
\resizebox{\linewidth}{!}{
\begin{tabular}{l | c c c c}
\hline
\textbf{Method} &
\textbf{Ax. FID ↓} &
\textbf{Sag. FID ↓} &
\textbf{Cor. FID ↓} &
\textbf{Avg. FID ↓} \\
\hline
HA-GAN              & 96.32 & 98.07 & 97.75 & 97.38 \\
3D-DDPM             & 79.53 & 78.16 & 76.07 & 77.92 \\
3D-LDM (w/ VQVAE)   & 67.17 & 63.43 & 62.96 & 64.52 \\
3D-LDM (w/ VAE)     & 40.69 & 37.34 & 37.83 & 38.62 \\
\textbf{3D-LLDM}    & \textbf{29.27} & \textbf{27.62} &
                      \textbf{28.04} & \textbf{28.31} \\
\hline
\end{tabular}}
\vspace{-3mm}
\caption{Comparison of FID scores (↓) among different generative models for medical image synthesis. 
Lower scores indicate higher image quality. 
The proposed 3D-LLDM model achieves the lowest FID scores in all views (Axial, Sagittal, and Coronal) as well as the overall average, consistently outperforming conventional generative models.}
\label{tab:fid_views}
\end{table}

\subsection{Evaluation of 3D-LLDM}
\vspace{-0.2cm}
This study evaluates the quality of generated images using the Fréchet Inception Distance (FID) metric. Feature extraction for FID computation was performed using a 3D ResNet-50 model from Tencent MedicalNet \cite{a3chen2019med3d},
which mapped $160 \times 160 \times 64$ volumetric images to 2048-dimensional feature vectors. 


Table~\ref{tab:fid_views} compares the proposed 3D-LLDM with HA-GAN, 3D-DDPM, and two variants of 3D latent diffusion models: (1) the default MONAI implementation and (2) a version with a VQ-VAE backbone. 
3D-LLDM achieves the lowest FID of \textbf{28.31}, representing a \textbf{70.9\%} relative reduction compared with HA-GAN (FID 97.38) and a \textbf{26.7\%} relative reduction compared with the strongest diffusion baseline (3D-LDM w/ VAE; FID 38.62).


\begin{table}[!t]
\centering
\tiny
\resizebox{\linewidth}{!}{%
\begin{tabular}{c | c c c c}
\hline
\textbf{CNN Model} & \textbf{Segmentation} & \textbf{R} & \textbf{R + S} & \textbf{Improvement} \\
\hline

\multirow{4}{*}{U-Net}
    & Liver-Only  & {\color{blue}\textbf{0.9650}} & \textbf{0.9662} & +0.124\% \\
    & Vein-Only   & 0.7905 & \textbf{0.8667} & +9.640\% \\
    & HCC-Only    & 0.7334 & \textbf{0.8152} & {\color{red}\textbf{+11.153\%}} \\
    & Multi-Class & 0.6968 & \textbf{0.7014} & +0.660\% \\
\hline

\multirow{4}{*}{ResUNet}
    & Liver-Only  & 0.9633 & \textbf{0.9634} & +0.010\% \\
    & Vein-Only   & 0.7197 & \textbf{0.7902} & +9.796\% \\
    & HCC-Only    & 0.7108 & \textbf{0.7646} & +7.569\% \\
    & Multi-Class & 0.6601 & \textbf{0.6652} & +0.773\% \\
\hline

\multirow{4}{*}{WideResUNet}
    & Liver-Only  & {\color{red}\textbf{0.9657}} & \textbf{0.9661} & +0.041\% \\
    & Vein-Only   & 0.7230 & \textbf{0.7989} & {\color{blue}\textbf{+10.498\%}} \\
    & HCC-Only    & 0.7251 & \textbf{0.7857} & +8.357\% \\
    & Multi-Class & 0.6961 & \textbf{0.7020} & +0.848\% \\
\hline

\multirow{4}{*}{DynUNet}
    & Liver-Only  & 0.9541 & {\color{red}\textbf{0.9736}} & +2.044\% \\
    & Vein-Only   & 0.6958 & \textbf{0.7540} & +8.365\% \\
    & HCC-Only    & 0.7002 & \textbf{0.7674} & +9.597\% \\
    & Multi-Class & 0.6983 & \textbf{0.7340} & +5.112\% \\
\hline

\multirow{4}{*}{VNet}
    & Liver-Only  & 0.9289 & {\color{blue}\textbf{0.9677}} & +4.177\% \\
    & Vein-Only   & 0.6908 & \textbf{0.7212} & +4.401\% \\
    & HCC-Only    & 0.6350 & \textbf{0.6825} & +7.480\% \\
    & Multi-Class & 0.5584 & \textbf{0.5937} & +6.322\% \\
\hline

\multirow{4}{*}{Overall Mean Dice}
    & Liver-Only  & 0.9544 & \textbf{0.9674} & +1.362\% \\
    & Vein-Only   & 0.7240 & \textbf{0.7862} & +8.591\% \\
    & HCC-Only    & 0.7009 & \textbf{0.7631} & +8.874\% \\
    & Multi-Class & 0.6619 & \textbf{0.6793} & +2.629\% \\
\hline

\end{tabular}
}
\vspace{-3mm}
\caption{
Comparison of segmentation performance using real data only (R) versus real data augmented with synthetic data (R + S) across different CNN architectures and segmentation tasks. The results demonstrate the effectiveness of synthetic data augmentation, with HCC-Only segmentation showing the highest individual improvement (U-Net: +11.153\%) and overall improvement (+8.874\% average), followed by Vein-Only segmentation (+8.591\% average). {\color{red}\textbf{Best}} and {\color{blue}\textbf{2nd-best}} results are highlighted.
}
\label{tab:segmentation}
\end{table}


Figure~\ref{fig:onecol} presents a qualitative comparison of synthetic MRI volumes generated by different models. The displayed images highlight structural variations in the liver and vascular structures. Compared to other models, 3D-LLDM generates more anatomically consistent images, accurately capturing finer details in liver margin sharpness and vascular structure delineation. This qualitative analysis makes 3D-LLDM well-suited for data augmentation in MR imaging tasks.

\begin{figure}[t]
\centering
\vspace{-2mm}
\makebox[\columnwidth]{%
    \includegraphics[width=1.01\columnwidth]{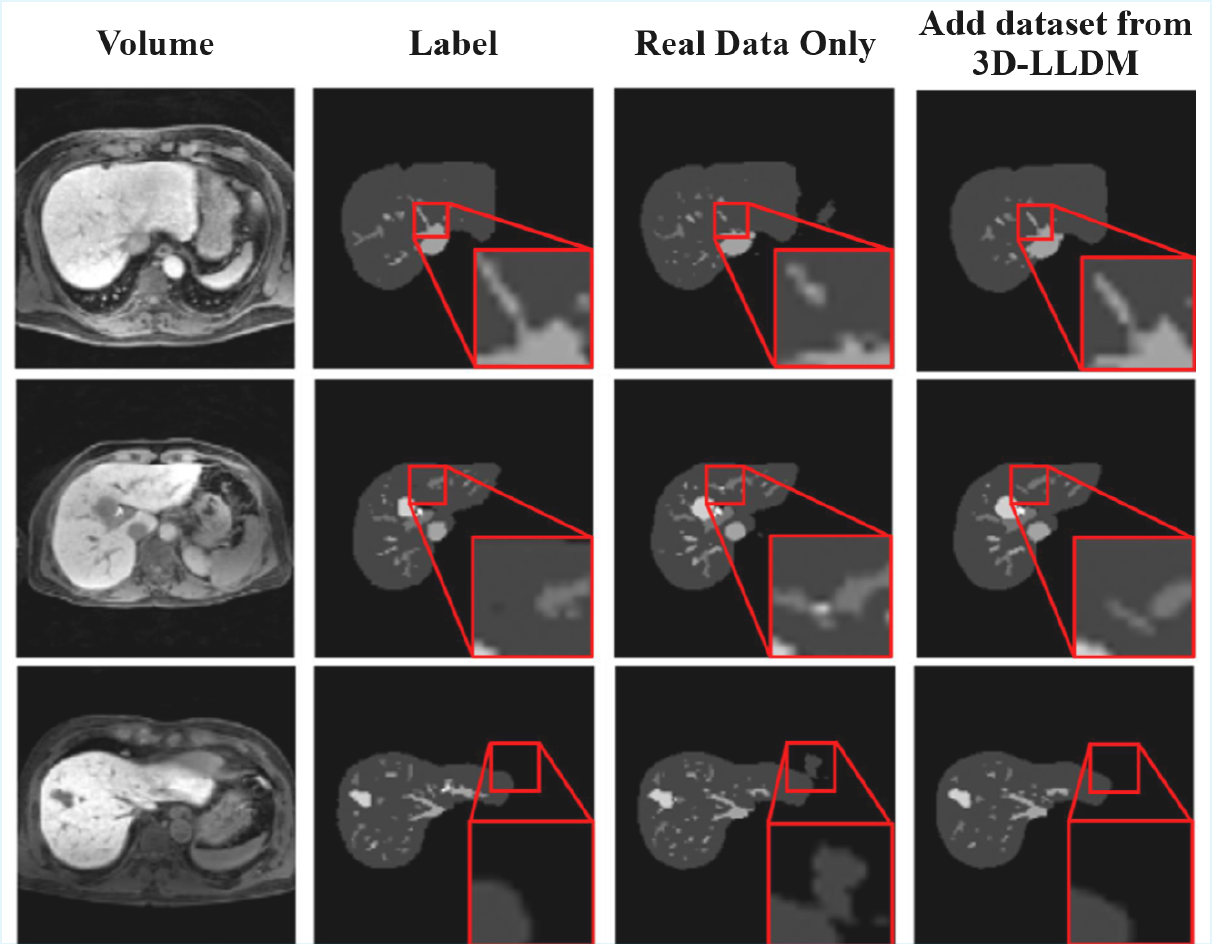}
}
\vspace{-6mm}
\caption{Qualitative comparison of multi-class segmentation results using U-Net with different training datasets. Each row represents a test sample, while the columns correspond to (1) input MR image, (2) ground truth segmentation, (3) segmentation using only real training data, and (4) segmentation using real+synthetic data from 3D-LLDM. Notably, the 2nd and 4th columns show a more continuous middle hepaticvein, whereas the 3rd column shows a discontinuous appearance.}
\label{fig:gen_visual}
\end{figure}

\subsection{Evaluation of Synthetic MR Volumes in Downstream Tasks}
\vspace{-0.2cm}
To evaluate the impact of 3D-LLDM-generated synthetic MR data on downstream tasks, we conducted comprehensive 3D segmentation experiments using real and synthetic label-volume pairs. We compared two training configurations: models trained exclusively on real data versus models trained on real data augmented with an equal proportion of synthetic samples.

The downstream 3D segmentation tasks encompassed four distinct scenarios: (1) hepatocellular carcinoma (HCC) tumor segmentation, (2) venous segmentation targeting portal and hepatic veins, (3) liver segmentation treating all anatomical structures except background as liver tissue, and (4) multi-class segmentation that simultaneously delineates portal veins, hepatic veins, and tumors as separate classes.

We evaluated five state-of-the-art CNN-based segmentation architectures: U-Net, ResUNet, WideResUNet, DynUNet, and VNet with increased channel depth. All models were implemented using the MONAI framework \cite{a2cardoso2022monai} and optimized using a combination of Dice loss and cross-entropy loss until validation loss convergence. Training employed the Adam optimizer with a learning rate of $10^{-4}$ and momentum of 0.95, requiring up to 18 hours per model on an NVIDIA A100 GPU.

Table~\ref{tab:segmentation} presents the segmentation performance comparison across different CNN architectures and training configurations. The incorporation of synthetic 3D-LLDM data consistently improved segmentation performance across most experimental settings. For liver segmentation, the average Dice score across all models reached 0.9674 with synthetic data augmentation, representing a 1.362\% improvement over real-data-only training. The most substantial improvements were observed in HCC segmentation, where synthetic data augmentation achieved an average improvement of 8.874\%, with U-Net showing the highest individual gain of 11.153\%. Vein segmentation also benefited significantly from synthetic data, with an average improvement of 8.591\%. These results demonstrate that label-guided synthesis particularly enhances performance for complex anatomical structures such as vessels and tumors, where data scarcity is a common challenge. 

Figure~\ref{fig:gen_visual} provides a qualitative comparison of the multi-class segmentation task with U-Net using different training datasets. In the fourth column, the segmentation model trained with synthetic data demonstrates improved accuracy, particularly in segmenting the middle hepatic vein, closely aligning with the ground truth labels shown in the second column. These findings validate that augmenting the training dataset with synthetic samples from 3D-LLDM enhances segmentation performance in downstream tasks.
\vspace{-0.1cm}
\section{Discussion and Conclusion}
\vspace{-0.1cm}
This study introduced the 3D-LLDM designed to generate high-resolution synthetic MR volumes along with the corresponding segmentation labels. Leveraging ControlNet for spatial guidance, our model ensured anatomically consistent synthesis and outperformed conventional generative models, achieving the lowest FID score and improving segmentation performance for the liver and multi-class tasks. Our study demonstrates the capability of 3D-LLDM to mitigate data scarcity in medical imaging, most notably in MR imaging. Subsequent work will aim to expand this approach across diverse imaging modalities and to exploit domain adaptation for improved generalization.
\vspace{-0.3cm}
\section{ACKNOWLEDGMENTS}
\vspace{-0.1cm}
\label{sec:acknowledgments}
This work was supported by the IITP grant funded by the Korea government (MSIT) under the AI Semiconductor Support Program (IITP-2023-RS-2023-00256081) and the Edge AI Semiconductor BMT Platform (No.\@~RS-2024-00399936).

\bibliographystyle{IEEEbib}

\end{document}